\documentclass[conference,letterpaper]{IEEEtran}

\usepackage[utf8]{inputenc} 
\usepackage[T1]{fontenc}
\usepackage{url}
\usepackage{ifthen}
\usepackage{cite}
\usepackage[cmex10]{amsmath} 

\usepackage{amssymb}
\usepackage{xcolor}
\usepackage{graphicx} 
\graphicspath{{figures/}}
\usepackage{booktabs}
\usepackage[font=footnotesize]{subcaption}
\usepackage{cleveref}

\newcommand{\figscale}{0.25}
\newcommand{\twocolfigscale}{0.45}

\begin{document}
\title{Correcting Class Imbalance in Prior-Data Fitted Networks for Tabular Classification} 

\author{
\IEEEauthorblockN{Samuel McDowell, Nathan Stromberg, and Lalitha Sankar}
\IEEEauthorblockA{
    School of Electrical, Computer and Energy Engineering\\
    Arizona State University,
    Tempe, AZ\\
    Email: \{scmcdowe, nstrombe, lsankar\}@asu.edu} 
}


\maketitle


\begin{abstract}
Prior-data fitted networks (PFNs) have achieved exceptional performance on tabular classification tasks. However, like other classifiers, their performance can suffer under the effect of class imbalance, resulting in poor performance for rare classes. 
Several techniques exist which attempt to mitigate the deleterious effect of class imbalance on classification performance, but the in-context learning (ICL) dynamic of PFNs means that loss-based strategies are impossible, and other techniques are unproven.
We have adapted several classical techniques addressing class imbalance and analyzed their performance on PFN classification. We observe that thresholding performs exceptionally well because of the calibration characteristics of PFNs, and downsampling performs comparably because of PFNs exceptional limited-data performance, with the additional benefit of reduced computation cost for inference.
\end{abstract}

\section{Introduction}
Prior-data fitted networks (PFNs) have become ubiquitous in critical areas such as tabular data inference \cite{grinsztajn2025tabpfn2.5} and timeseries forecasting \cite{onur2025timepfn}. Specifically, TabPFN (and its variants) has demonstrated extraordinary performance on tabular classification and regression tasks \emph{without updating any weights} and with very little task-specific training data. PFNs achieve this outstanding efficiency by pretraining on copious amounts of \emph{synthetic} data specifically for the task of in-context learning. 

By training on synthetic data from only highly structured causal models, PFNs implicitly learn to estimate the posterior predictive distribution (PPD) over a prior determined by the class of data generating models. The relationship between the pretraining model class and the final prediction is critical,
as the predictions of the classifier will follow the structure of the data generating models used in training.
In this way, controlling the pretraining prior controls the hypothesis space of the classification.


Another key architectural aspect of PFNs is their reliance on in-context learning (ICL). ICL is a form of meta-learning where the PFN model is pretrained not to predict outputs for a fixed task, but to predict the relationship between a labeled context set and an unlabeled query. This results in a model whose weights are not updated at all for a new task; instead, the prediction relies on a relatively few labeled examples (context) given to the model. This is distinct from classical in-weight learning where the task-specific examples are used to update model weights. 
ICL is typically performed using transformer models, so the relationships between context and query is captured using the attention mechanism. 

While PFNs achieve state of the art performance on tabular classification, like most classification models, they perform poorly on class-imbalanced data, where some classes are vastly more prevalent than others. In fact, these models may achieve reasonable or even stellar average accuracy, but the performance on the minority classes can be significantly degraded due to limited samples. This, in turn, limits detection of rare classes (e.g., rare disease or cyberattack detection).

Addressing the effects of class imbalance falls in three major categories: loss-, data-, or decision-based methods. Loss reweighting is a classical method that falls in the first category, and involves upweighting the loss on minority class samples such that they have the same influence on the model as the majority. Data-level methods include downsampling the majority size to match that of the minority as well as generating synthetic minority samples. Decision-level methods involve manipulating the output of the model, such as scaling/tilting the soft score from the classifier. All of these methods have merit; however, since the learning dynamics of ICL are distinct from in-weight learning, some of these techniques, in particular loss reweighting, cannot be applied to this setting. 

Recently, \cite{ma2024tabpfgen} addressed the performance of TabPFNs on rare classes by evaluating the efficacy of generating synthetic minority samples. This technique is limited by the effectiveness of the method used to generate the synthetic samples, as any distortion of the distribution of the synthetic samples will affect the downstream classification. It is also computationally expensive, since a sample must be generated for each additional sample in the overrepresented classes.

We begin by investigating the unique calibration characteristics of PFNs. This motivates our empirical evaluation of theoretically-grounded and practically useful correction methods including 
thresholding, downsampling, oversampling, and synthetic upsampling (all defined formally in \Cref{sec:data_level_strategies}). For binary classification tasks, our theory-guided experimental results show that thresholding achieves the best performance, with a drastic increase in minority class performance with a minimal decrease in majority performance. Downsampling also performs well, achieving the highest worst-class accuracy with only a slight decrease in balanced performance and the additional benefit of decreasing inference computation cost by reducing the number of context samples.

\section{Problem Setup}

\subsection{Prior-Data Fitted Networks (PFNs)}
PFNs are a model class trained on a Bayesian prior over supervised learning tasks, typically with large-scale synthetic datasets, so that they can use in-context learning (ICL) to predict the posterior predictive distribution (PPD) directly \cite{muller2021pfns}. During training, these models are given a set of context points drawn from a distribution and trained to predict the masked label of a query point drawn from the same distribution. Minimizing a standard cross entropy loss over a variety of distributions leads the model to learn $P(y \mid x, D)$, the PPD of the context $D=\{x_c,y_c\}_{i=1}^n$ and query $x$. Formally, the PPD can be written as \cite{hollmann2022tabpfn}:

\begin{equation}
P(y|x, D) \propto \int_{\Phi} P(y|x, \phi)P(D|\phi)P(\phi)d\phi,
\end{equation}
where $y$ is the class label of data point $x$, and $D$ is a labeled dataset drawn from the same distribution as $x$, and $\Phi$ is the set of data generating functions.

The set of data generating functions is a prior which defines the hypothesis space of the classifier. A common selection is the set of structural causal models (SCMs), where the causal relationships between features are represented by the edges of a directed acyclic graph~\cite{hollmann2022tabpfn}. This trains the model to estimate the PPD directly by \emph{implicitly} predicting the SCM which best fits the data and using that model to predict $\hat{y}$.


\subsection{In-Context Learning (ICL)}

During ICL, the pretrained model is given a set of $n$ labeled context points, $\{(x_c, y_c)\}_{i=1}^n$, and a query, $x_q$, and is asked to predict $\hat{y}_q$, the label of the query point. Critically, this does not involve updating model weights and is thus distinct from standard in-weight learning, where labeled data is used to update model weights before passing in the query to predict a label.

The idea of ICL originated with large languages models (LLMs), when their ability to perform task-agnostic few-shot classification~\cite{brown2020llmicl}. However, since then, the idea of ICL has expanded, and models have been trained to perform ICL explicitly in a variety of settings, including images~\cite{wang2023imageicl} and tabular data~\cite{hollmann2022tabpfn}.

The learning of an ICL model takes place entirely within the latent representation of the context and query data rather than the weights of the model, which can make manipulating the model difficult. Without performing a computationally expensive fine-tuning operation on the model, the only ways to affect its performance are to manipulate the input data or the downstream prediction.

\subsection{Receiver Operating Characteristic (ROC)}

For binary classification with classes `0' and `1' (minority class), the ROC curve defines the possible operating points of a classifier with certain misdetection (MD) and false alarm (FA) rates. A perfect classifier will have MD and FA rate of 0; that is, $P(\hat{y}=0 \mid y=1) =0$ and $P(\hat{y}=1 \mid y=0) =0$. 

In practice,  there is a tradeoff between MD and FA rates: as MD decreases, FA increases and vice versa. Depending on the cost of misdetections and false alarms, an operating point for the classifier can be selected to minimize expected cost. One common choice is to minimize the probability of error:
\begin{equation}
P_e = \pi_1 P(\hat{y}=0 \mid y=1) + \pi_0 P(\hat{y}=1 \mid y=0),
\label{eqn:prob_error}
\end{equation}
where $\pi_0$ and $\pi_1$ are the prior probabilities of classes `0' and `1', respectively. In fact, this is what will be achieved by minimizing an unweighted loss, achieved in practice using empirical risk minimization (ERM). 

This, however, does not necessarily result in a good downstream classifier. When performing ERM over an imbalanced dataset, a classifier will reach an operating point with a high misdetection or false alarm rate. 
Hereafter, we assume that `0' and `1' are the majority and minority classes, respectively. 



\subsection{Calibration}

Let $f_\theta(x)$ be a pretrained classifier, parametrized by $\theta$. We view the classifier output as a soft score, i.e., $f_\theta(x) \in [0,1]$.
A model is \emph{perfectly calibrated} if $P[Y=y|f_{\theta}(X) = p] = p, \text{ for all } p\in [0,1]$. That is, the predicted probability that sample $x$ has label $y$ is the same as the true posterior. Since $Y$ is a Bernoulli RV, the probability is simply the expectation (or empirically, the sample average).

\begin{figure}[htbp]
  \centering
  \begin{subfigure}[b]{\twocolfigscale\columnwidth}
    \centering
    \includegraphics[width=\textwidth]{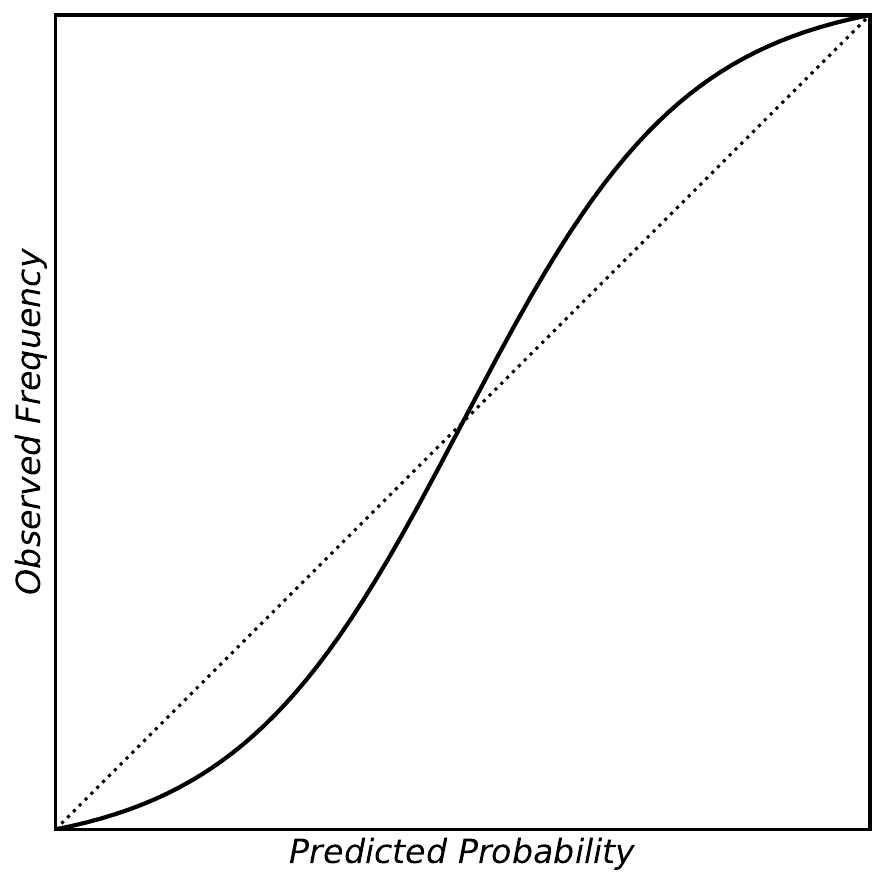}
    \caption{Underconfident}
    \label{fig:underconfident}
  \end{subfigure}
  \hfill
  \begin{subfigure}[b]{\twocolfigscale\columnwidth}
    \centering
    \includegraphics[width=\textwidth]{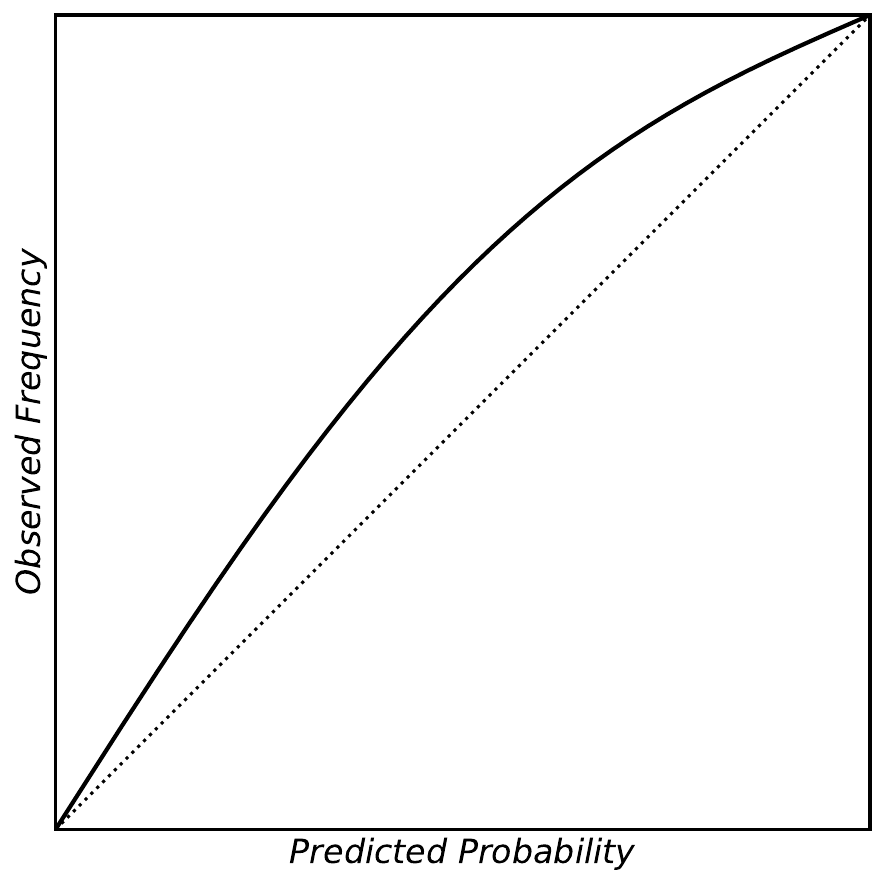}
    \caption{Class 0 Biased}
    \label{fig:0_biased}
  \end{subfigure}
  \caption{Example Calibration Curves}
  \label{fig:ex_calibration}
\end{figure}

A standard method to evaluate calibration of a model is to plot the \emph{observed frequency} of $p \in [0,1]$ as a function of the  \emph{predicted probability}, i.e., $\mathbb{E}[y \mid f_{\theta}(x) = p]$. In practice, the x-axis values are obtained using empirical estimates for a chosen range of $p$. And the y-axis values are the empirical frequency of the label in the same range. A perfectly calibrated curve is a straight line running through the origin. A model is \emph{underconfident} (resp. \emph{overconfident}) if the predicted probability is biased towards (resp. away from) 0.5 than the true probability for all $p$. 
On the other hand, a model that consistently predicts above or below the observed frequency is biased towards one class or the other (see \Cref{fig:0_biased}). Calibration information allows making corrections to $f_\theta(x)$ to make more accurate downstream predictions.

%

\subsection{Data-Level Strategies (Sampling)}
\label{sec:data_level_strategies}

\noindent\textbf{Downsampling} involves removing majority samples from the context set, such that the number of samples from each class becomes equal. This achieves $\pi_0=\pi_1$ at the cost of decreasing the available information about the majority class.

\noindent\textbf{Oversampling} involves including samples from the minority class in the context set multiple times to equalize the number of samples from each class. This technique also achieves $\pi_0=\pi_1$, but distorts the minority distribution, making it appear spikier than the true distribution.

\noindent\textbf{Synthetic Upsampling} involves generating artificial samples of the minority class using the context set and supplementing the context set with enough of these samples that it becomes class-balanced. Synthetic upsampling is similar to oversampling, but the distortion of the minority distribution comes from any inaccuracies in the distribution learned by the generator.

\subsection{Decision-Level Strategies}
The Bayes optimal rule for classification picks the class with the highest soft-score. For binary classification, this simplifies to setting a threshold of 0.5 to make a hard decision. 

\textbf{Thresholding} involves moving the decision boundary away from 0.5.
This can be interpreted as minimizing risk $\mathcal{R}$ when the cost of false negatives (misclassifying a minority as a majority) and false positives (misclassifying a majority as a minority) are unequal, i.e.,
\begin{equation}
    \mathcal{R}=C_{01}\pi_1 P(\hat{y}=0 \mid y=1) + C_{10}\pi_0 P(\hat{y}=1 \mid y=0).
\end{equation}

If we have a classifier optimized with $\pi_0 \neq \pi_1$, we can adapt it to be equivalent to the classifier which would be reached with $\pi_0=\pi_1$ by defining $C_{01}/C_{10}=\pi_0/\pi_1$. Determining a classification threshold, $\tau$, so that our result is $1[f_\theta(x,D) > \tau]$, we have 
\begin{equation}
    \tau=\frac{C_{10}}{C_{10}+C_{01}}=\pi_1.
\end{equation}
Thus, by adjusting the classification threshold away from 0.5, we are able to counteract the effect of data imbalance.

\subsection{Metrics}

Our metrics of interest are the per-class, balanced, and worst-class accuracy (WCA). We use a hold-out test set to compute accuracies. Per-class accuracy is simply the fraction of correctly classified points in each class of the test data, i.e., $P(\hat{y}=i|y=i), i\in\{0,1\}$. The average test accuracy is simply the empirical accuracy over both classes, i.e., each accuracy is scaled by its empirical prior. On the other hand, balanced accuracy is the average of the two per-class accuracies (equivalently, $1-P_e$, where $P_e$ is defined in \Cref{eqn:prob_error} with $\pi_0=\pi_1$). Finally, WCA is the minimum of the two per-class accuracies.

\section{Experimental Results \& Analysis}

For our experiments, we select binary classification tasks from the benchmark collection of datasets, OpenML-CC18. This collection has 72 tabular classification tasks from which we select 11 datasets satisfying the following constraints:
\begin{itemize}
    \item \textbf{Task}: binary classification
    \item \textbf{Test Size}: 500 examples per class
    \item \textbf{Train Size}: 500 minority examples and 950 majority examples (allowing imbalances of up to $\pi_1 = 0.05$ while maintaining 1000 total training examples)
\end{itemize}
The query (test) set is always balanced, and thus, the average test accuracy can also be viewed as the {balanced accuracy}. 

In these experiments, we utilize TabPFN-2.5~\cite{grinsztajn2025tabpfn2.5} which offers state-of-the-art performance among tabular PFN models. We list the full selection of datasets used in \Cref{tab:datasets} along with their total sample counts and natural imbalance.

\begin{table}[htbp]
\centering
\caption{Summary of the datasets used in experiments.}
\label{tab:datasets}
\begin{tabular}{lrr}
\toprule
Dataset & $N$ & $\pi_1$ \\
\midrule
\texttt{kr-vs-kp} & 3,196 & 0.478 \\
\texttt{spambase} & 4,601 & 0.394 \\
\texttt{electricity} & 45,312 & 0.424 \\
\texttt{jm1} & 10,885 & 0.194 \\
\texttt{adult} & 48,842 & 0.239 \\
\texttt{Bioresponse} & 3,751 & 0.458 \\
\texttt{phoneme} & 5,404 & 0.293 \\
\texttt{nomao} & 34,465 & 0.286 \\
\texttt{PhishingWebsites} & 11,055 & 0.443 \\
\texttt{bank-marketing} & 45,211 & 0.117 \\
\texttt{numerai28.6} & 96,320 & 0.495 \\
\bottomrule
\end{tabular}
\end{table}

\subsection{Calibration}

\begin{figure}[htbp]
  \vspace{1mm}
  \centering
  \includegraphics[width=\figscale\textwidth]{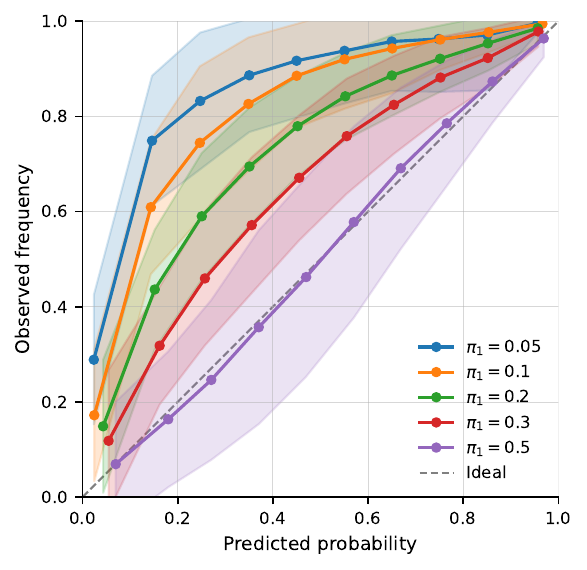}
  \caption{Calibration Curve Averaged Over Datasets and $N \in \{100, 500, 1000\}$}
  \label{fig:avg_calibration}
\end{figure}

In \Cref{fig:avg_calibration}, we illustrate the empirically computed calibration of TabPFN over several datasets when the training context is either balanced or imbalanced. 
We observe a persistent trend of good calibration with balanced datasets and majority class bias in the imbalanced setting.
\Cref{fig:avg_calibration} shows the calibration curve averaged over all of the datasets and three context sizes. In the balanced setting ($\pi_1=0.5$), we see that the model is well-calibrated, but as the data becomes imbalanced, the model becomes increasingly majority-biased. 
In general, most non-linear classifier models are not well-calibrated and are either overconfident or underconfident for all predicted probabilities, and are often addressed with tilting or scaling methods \cite{niculescu-mizil_predicting_2005}. 
In contrast, TabPFNs are simply biased towards the majority class for which the solution can be simple thresholding. This is a critical observation which, to the best of our knowledge, has not been exploited in the PFN literature.

\subsection{ROC}
\begin{figure}[htbp]
  \centering
  \includegraphics[width=\figscale\textwidth]{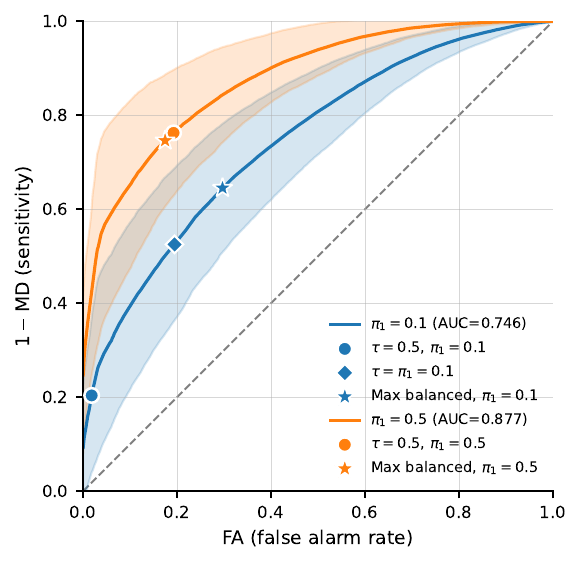}
  \caption{ROC Curve for \texttt{kr-vs-kp} with $N=25$}
  \label{fig:task3_roc}
\end{figure}

While we see that TabPFN is uncalibrated in general for the standard threshold, the effect of other thresholds is unclear. We compute an estimate of the ROC curve of TabPFN-2.5 for each chosen imbalance level. \Cref{fig:task3_roc} shows the ROC curve for the \texttt{kr-vs-kp} dataset with $N=25$ samples and $\pi_1 \in \{0.1, 0.5\}$. On each curve, we also highlight the $(\text{FA}, \text{MD})$ choice for $\tau=0.5$ as a round filled marker and that for $\tau^*$ that achieves the maximum balanced accuracy as a diamond filled marker.  We see that the AUC falls significantly with increasing levels of imbalance. Furthermore, without threshold correction, i.e., for threshold $\tau=0.5$, for $\pi_1=0.1$, the model reaches an operating point with a very high error rate for the minority class ($\text{MD}\approx0.8$). By adjusting the threshold to $\tau=\pi_1$, we can push the operating point towards the maximal balanced accuracy point, indicated by blue filled star. Note that the $\tau^*$ (indicated by the red filled star) for the $\pi_1=0.5$ curve is close to the ideal $0.5$ threshold point but not exactly the same due to empirical and data size limited computations.

\subsection{Thresholding}
\begin{figure}[htbp]
    \vspace{1mm}
    \centering
    \begin{subfigure}[b]{\twocolfigscale\columnwidth}
        \centering
        \includegraphics[width=\textwidth]{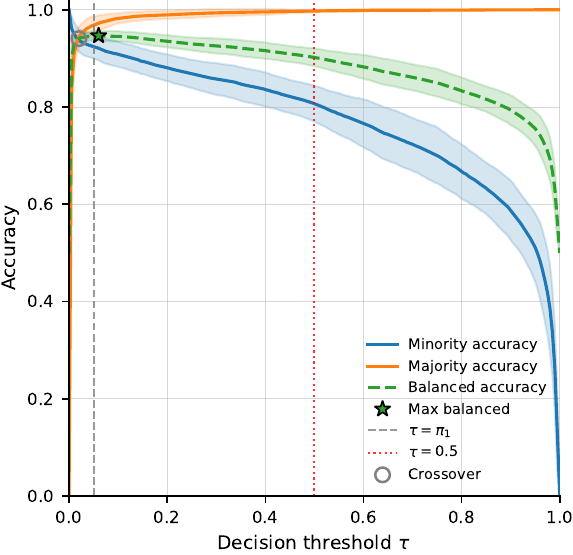}
        \caption{$\pi_1 = 0.05$}
        \label{fig:sub_a}
    \end{subfigure}
    \hfill
    \begin{subfigure}[b]{\twocolfigscale\columnwidth}
        \centering
        \includegraphics[width=\textwidth]{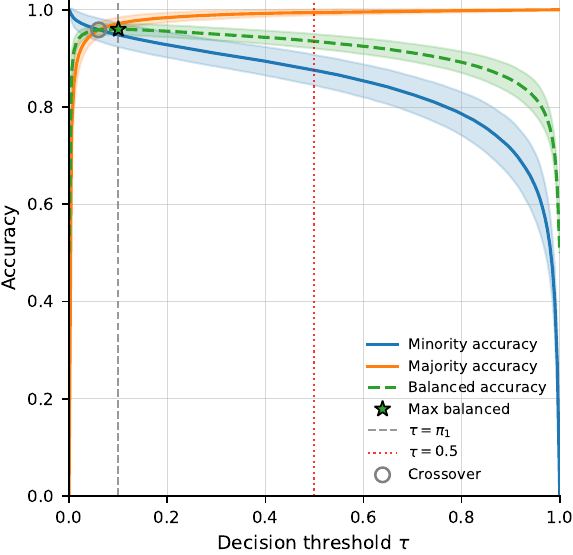}
        \caption{$\pi_1 = 0.1$}
        \label{fig:sub_b}
    \end{subfigure}
    \begin{subfigure}[b]{\twocolfigscale\columnwidth}
        \centering
        \includegraphics[width=\textwidth]{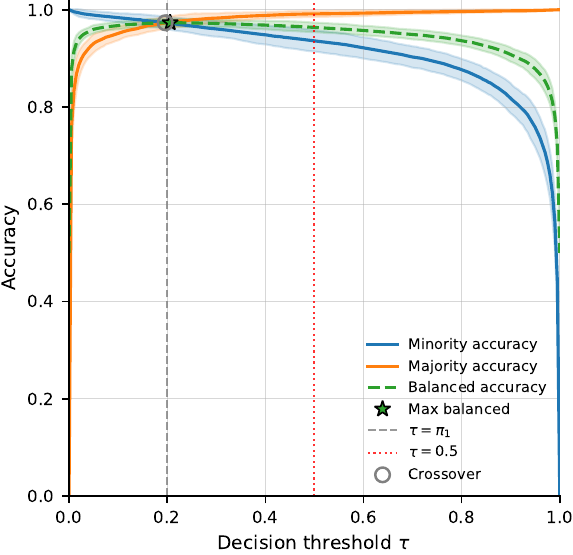}
        \caption{$\pi_1 = 0.2$}
        \label{fig:sub_c}
    \end{subfigure}
    \hfill
    \begin{subfigure}[b]{\twocolfigscale\columnwidth}
        \centering
        \includegraphics[width=\textwidth]{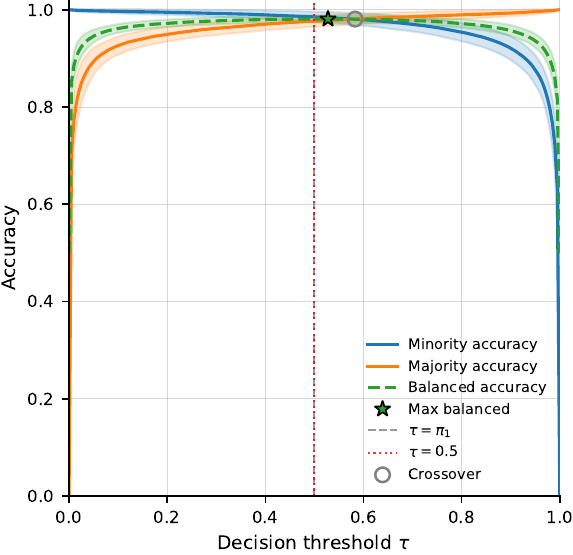}
        \caption{$\pi_1 = 0.5$}
        \label{fig:sub_e}
    \end{subfigure}
    \caption{Threshold crossover for \texttt{kr-vs-kp} with $N=500$ across imbalance levels $\pi_1 \in \{0.05, 0.1, 0.2, 0.5\}$}
    \label{fig:task3_threshold_crossover}
\end{figure}

Based on the observations on the ROC curve, it is natural to ask how sensitive the per-class accuracies are to the choice of threshold for a given dataset. To evaluate this effect, we evaluate TabPFN performance with a variety of decision thresholds, $\tau$. When we do this, we observe that the maximum balanced accuracy appears approximately at the value $\tau=\pi_1$, aligning with the $P_e$ minimization calculation.
\Cref{fig:task3_threshold_crossover} shows an example of one such experiment on the \texttt{kr-vs-kp} dataset with a context size of $N=500$ samples and an imbalance of $\pi_1 \in \{0.05, 0.1, 0.2, 0.5\}$. We see that even as the imbalance changes and the crossover point moves, the maximum balanced accuracy point tracks $\tau=\pi_1$.

This result validates our decision to use a threshold of $\tau=\pi_1$ as a correction method to attempt to maximize balanced accuracy. We remark that the idea of thresholding is very standard in detection theory; however, in modern learning-based classification, average accuracy (averaged over priors) is often the only metric presented which results in these models ignoring the performance on rare classes, especially when the average accuracy is high. Our work shows that the performance of SOTA models like TabPFN can be enhanced by judicious use of thresholding and downsampling.

\subsection{Downsampling}

While thresholding attempts to adjust the output of the model to account for a difference in the priors, downsampling adjusts those priors directly by reducing the number of majority samples. To investigate the effect of different levels of downsampling on classification performance, we fix a number of class 1 (minority) samples and sweep the number of class 0 (majority) samples. Note that we still denote class 0 as the majority, even when it is sampled down such that $\pi_0 < \pi_1$.

\begin{figure}[htbp]
  \centering
  \includegraphics[width=0.4\textwidth]{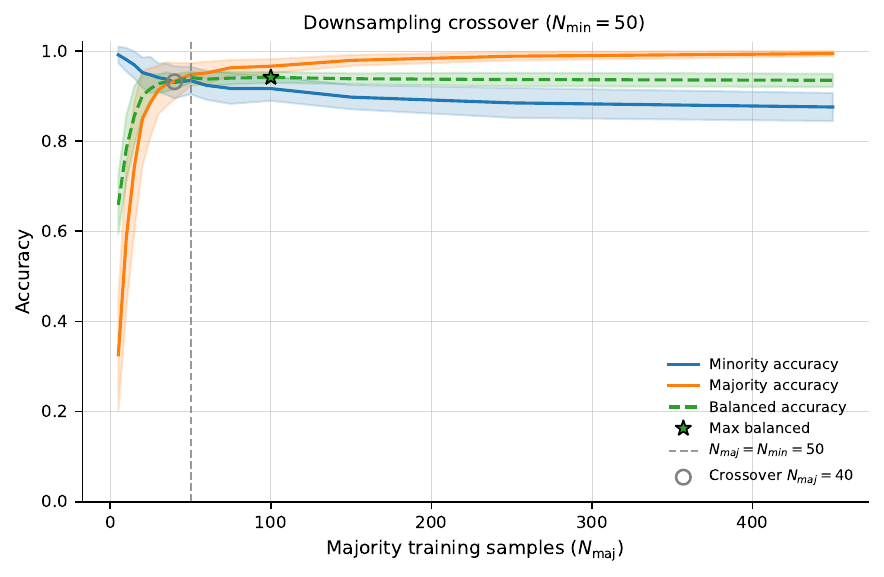}
  \caption{TabPFN2.5 Downsample Crossover}
  \label{fig:task3_downsample_crossover}
\end{figure}

In these experiments, we see that the balanced accuracy is approximately constant while the number of class 0 samples is greater than or equal to the number of class 1 samples, though as the number of class 0 samples increases the worst-class accuracy decreases. \Cref{fig:task3_downsample_crossover} shows an example of this evaluation for \texttt{kr-vs-kp} with $N_{min}=50$.

This justifies the decision to downsample to $\pi_0=\pi_1$. Downsampling to the lowest level before the rapid falloff in balanced accuracy has the additional benefit of noticeably reducing required computation, since TabPFN query computation scales quadratically with context size~\cite{hollmann2022tabpfn}.

\begin{table*}[htbp]
\vspace{2mm}
\centering
\caption{Results by Dataset and Correction Method Averaged Over $N \in \{100, 500, 1000\}$ and $\pi_1 \in \{0.05, 0.1, 0.2, 0.3\}$}
\label{tab:accuracy}
\resizebox{0.75\textwidth}{!}{%
\begin{tabular}{l@{\hspace{6pt}}rr@{\hspace{6pt}}rr@{\hspace{6pt}}rr@{\hspace{6pt}}rr@{\hspace{6pt}}rr}
\toprule
Dataset & \multicolumn{2}{c}{None} & \multicolumn{2}{c}{Thrsh.} & \multicolumn{2}{c}{OS} & \multicolumn{2}{c}{TabPFGen} & \multicolumn{2}{c}{DS} \\
\cmidrule(lr){2-3}\cmidrule(lr){4-5}\cmidrule(lr){6-7}\cmidrule(lr){8-9}\cmidrule(lr){10-11}
 & Bal. & WCA & Bal. & WCA & Bal. & WCA & Bal. & WCA & Bal. & WCA \\
\midrule
\texttt{kr-vs-kp} & 0.931 & 0.868 & \textbf{0.956} & \textbf{0.938} & 0.783 & 0.569 & 0.912 & 0.831 & 0.928 & 0.902 \\
\texttt{spambase} & 0.866 & 0.753 & \textbf{0.920} & \textbf{0.901} & 0.700 & 0.408 & 0.789 & 0.593 & 0.903 & 0.876 \\
\texttt{electricity} & 0.645 & 0.317 & \textbf{0.757} & \textbf{0.678} & 0.574 & 0.168 & 0.638 & 0.305 & 0.734 & 0.677 \\
\texttt{jm1} & 0.534 & 0.096 & \textbf{0.637} & 0.537 & 0.529 & 0.102 & 0.539 & 0.113 & 0.632 & \textbf{0.560} \\
\texttt{adult} & 0.665 & 0.372 & \textbf{0.795} & \textbf{0.752} & 0.578 & 0.179 & 0.664 & 0.371 & 0.783 & 0.726 \\
\texttt{Bioresponse} & 0.588 & 0.212 & \textbf{0.704} & \textbf{0.646} & 0.539 & 0.094 & 0.585 & 0.204 & 0.683 & 0.643 \\
\texttt{phoneme} & 0.666 & 0.379 & \textbf{0.811} & \textbf{0.770} & 0.602 & 0.230 & 0.629 & 0.292 & 0.791 & 0.741 \\
\texttt{nomao} & 0.867 & 0.761 & \textbf{0.915} & \textbf{0.897} & 0.715 & 0.438 & 0.821 & 0.660 & 0.905 & 0.884 \\
\texttt{PhishingWebsites} & 0.882 & 0.779 & \textbf{0.921} & \textbf{0.893} & 0.758 & 0.525 & 0.872 & 0.757 & 0.913 & 0.889 \\
\texttt{bank-marketing} & 0.664 & 0.373 & \textbf{0.808} & \textbf{0.774} & 0.555 & 0.126 & 0.652 & 0.346 & 0.789 & 0.753 \\
\texttt{numerai28.6} & 0.500 & 0.001 & \textbf{0.504} & 0.235 & 0.500 & 0.007 & 0.500 & 0.002 & 0.503 & \textbf{0.452} \\
\midrule
Average & 0.710 & 0.446 & \textbf{0.793} & 0.729 & 0.621 & 0.259 & 0.691 & 0.407 & 0.779 & \textbf{0.737} \\
\bottomrule
\end{tabular}
}
\end{table*}

\subsection{Results}

\Cref{tab:accuracy} shows the results of each correction method for each dataset tested. For each correction method, we report the balanced, and worst-class accuracy.

We see that for most datasets, thresholding achieves both strong balanced and worst-class accuracy, with downsampling close behind. Meanwhile, both OS and TabPFGen perform worse than the base model alone. For OS, this can be attributed to the PFN fitting to the repeated samples resulting in a spiky posterior fit. Meanwhile, it is unsurprising that minority points generated by TabPFGen would not improve the accuracy of the TabPFN classifier given that they are from the same model class and using the same data.

\Cref{fig:method_rel_accuracy} shows the change in accuracy for each method relative to no correction. We see that thresholding and downsampling both increase minority accuracy significantly at the cost of a slight decrease in majority accuracy. Downsampling results in a slightly greater increase in minority class accuracy, but a much greater decrease in majority accuracy, leading thresholding to have a slightly higher balanced accuracy.

\newcommand{\deltaplotscale}{0.49}
\begin{figure}[htbp]
    \centering
    \begin{subfigure}[b]{\deltaplotscale\columnwidth}
        \centering
        \includegraphics[width=\textwidth]{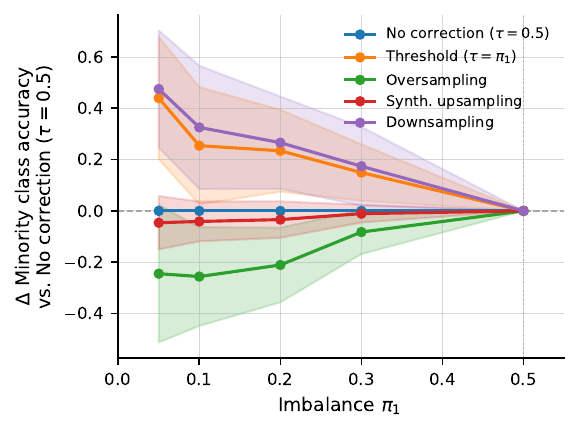}
        \caption{$\Delta$ Minority}
        \label{fig:rel_min}
    \end{subfigure}
    \begin{subfigure}[b]{\deltaplotscale\columnwidth}
        \centering
        \includegraphics[width=\textwidth]{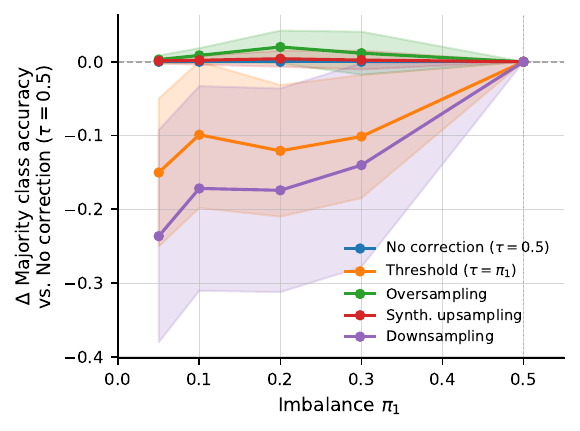}
        \caption{$\Delta$ Majority}
        \label{fig:rel_maj}
    \end{subfigure}
    \begin{subfigure}[b]{\deltaplotscale\columnwidth}
        \centering
        \includegraphics[width=\textwidth]{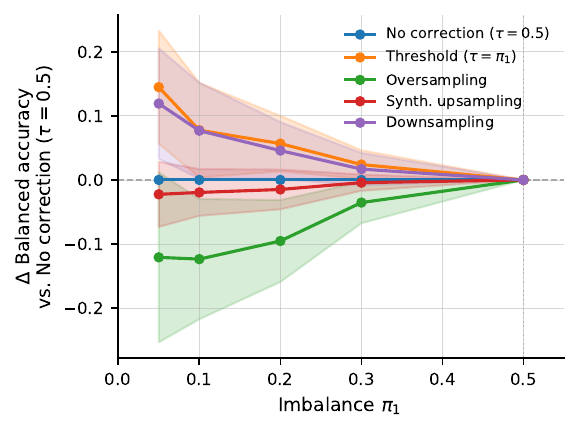}
        \caption{$\Delta$ Balanced}
        \label{fig:rel_balanced}
    \end{subfigure}
    \begin{subfigure}[b]{\deltaplotscale\columnwidth}
        \centering
        \includegraphics[width=\textwidth]{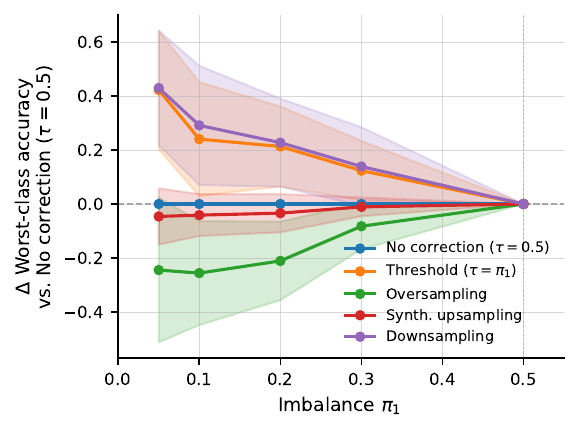}
        \caption{$\Delta$ WCA}
        \label{fig:rel_wca}
    \end{subfigure}
    \caption{Change ($\Delta$ \emph{relative to no correction}) in minority \eqref{fig:rel_min}, majority \eqref{fig:rel_maj}, balanced \eqref{fig:rel_balanced}, and worst-class accuracy \eqref{fig:rel_wca} averaged over datasets and $N \in \{100, 500, 1000\}$}
    \label{fig:method_rel_accuracy}
\end{figure}

\section{Conclusion}

Through a thorough empirical evaluation of PFNs on imbalanced tabular classification tasks, we elucidate both a key failure mode of these powerful models and a suggestion as to a solution. We test both contemporary and classical methods which are amenable to the ICL framework and find that simple thresholding outperforms more complex corrections. We observe that the methods which perform the best are simple and easy to implement, meaning practitioners will be able to effectively address imbalance and achieve high accuracy on classes with limited data without the addition of significant complexity. However, this simplicity may also leave some performance to be gained through methods developed specifically for the strengths of PFNs. Future work will extend this framework to multi-class classification, which will require non-trivial adaptations of the correction strategies explored here. More broadly, we hope this work opens a productive line of inquiry into principled methods for classification bias correction, especially for in-context learning in foundation models.

\IEEEtriggeratref{6}
\bibliographystyle{IEEEtran}
\bibliography{references}

\end{document}